\documentclass[letterpaper, 10 pt, conference]{ieeeconf}  

\usepackage[utf8]{inputenc}
\usepackage{graphicx}
\usepackage{epstopdf}
\usepackage{etoolbox}
\usepackage{cite}
\usepackage{picinpar}
\usepackage{amsmath}
\usepackage{url}
\usepackage{flushend}
\usepackage{textcomp}
\usepackage{bm}
\usepackage{xcolor}
\usepackage{textcomp}
\usepackage{comment}
\usepackage{makecell}
\usepackage{amsmath,amsfonts,mathrsfs}
\usepackage{amsmath}     

\IEEEoverridecommandlockouts                              

\title{\LARGE \bf
Development of a Multi-Fingered Soft Gripper Digital Twin\\ for Machine Learning-based Underactuated Control
}

\author{Wu-Te Yang$^{1}$, Pei-Chun Lin$^{2}$
\thanks{$^{1}$The author is an independent researcher with PhD from the University of California, Berkeley, USA
        {\tt\small wtyang@berkeley.edu}}%
\thanks{$^{2}$The author is with the Department of Mechanical Engineering,
        National Taiwan University, Bio-inspired Robotics Lab, Taipei, Taiwan
        {\tt\small peichunlin@ntu.edu.tw}}%
}

\begin{document}

\maketitle
\thispagestyle{empty}
\pagestyle{empty}

\begin{abstract}

Soft robots, made from compliant materials, exhibit complex dynamics due to their flexibility and high degrees of freedom. Controlling soft robots presents significant challenges, particularly underactuation, where the number of inputs is fewer than the degrees of freedom. This research aims to develop a digital twin for multi-fingered soft grippers to advance the development of underactuation algorithms. The digital twin is designed to capture key effects observed in soft robots, such as nonlinearity, hysteresis, uncertainty, and time-varying phenomena, ensuring it closely replicates the behavior of a real-world soft gripper. Uncertainty is simulated using the Monte Carlo method. With the digital twin, a Q-learning algorithm is preliminarily applied to identify the optimal motion speed that minimizes uncertainty caused by the soft robots. Underactuated motions are successfully simulated within this environment. This digital twin paves the way for advanced machine learning algorithm training.

\end{abstract}

\section{INTRODUCTION}
\label{introduction}
Soft robotics is a rapidly growing field that provides solutions for tasks where traditional robots face limitations. Specifically, the flexibility and adaptability of soft actuators offer distinct advantages for applications requiring careful manipulation~\cite{navas2021gripper, george2020survey} and interaction with complex or unpredictable environments~\cite{fumiya2011review}. In contrast to rigid robotic hands, soft grippers excel at conforming to various object shapes and sizes, making them valuable in areas like medical robotics~\cite{alici2018bending} and human-robot interaction~\cite{demir2020design}. However, controlling the motion and coordination of multiple soft fingers in a gripper remains a challenge~\cite{yang2023control}. Meanwhile, as underactuated control reduces the number of inputs and optimizes the dimensions, developing effective underactuated control techniques remains an open challenge.~\cite{daniela2023survey}. 

Underactuation describes a system that has fewer actuators than degrees of freedom~\cite{underactuated2019he, tuna2019syn}. Due to the fewer actuators, underactuated systems have limited control over all degrees of freedom. Achieving stability and robust control is much harder in underactuated systems because of the same reason~\cite{underactuated2019he}. Systems with uncertainty such as soft robots may exacerbate this issue. However, underactuation for soft robots is achievable and applicable under some conditions and assumptions~\cite{yang2024control}. When soft robotic systems fail to satisfy these conditions or assumptions, the underactuation strategy may not be effective. Machine learning has gained significant popularity recently, and machine learning-based approaches such as reinforcement learning could offer a viable alternative~\cite{russel2016ai}. Reinforcement learning (RL) is a technique that enables an agent to learn how to make decisions in order to achieve the best possible outcome~\cite{sutton2018rl}. Simulation environments are commonly used to train RL algorithms, facilitating sim-to-real skill transfer~\cite{zhao2020simtoreal}. However, a challenge arises from the complexity and unpredictability of soft materials in simulation environments for soft robots. These properties can make the simulation results difficult to reference.

\begin{figure}[t]
    \centering
    \includegraphics[width=210pt]{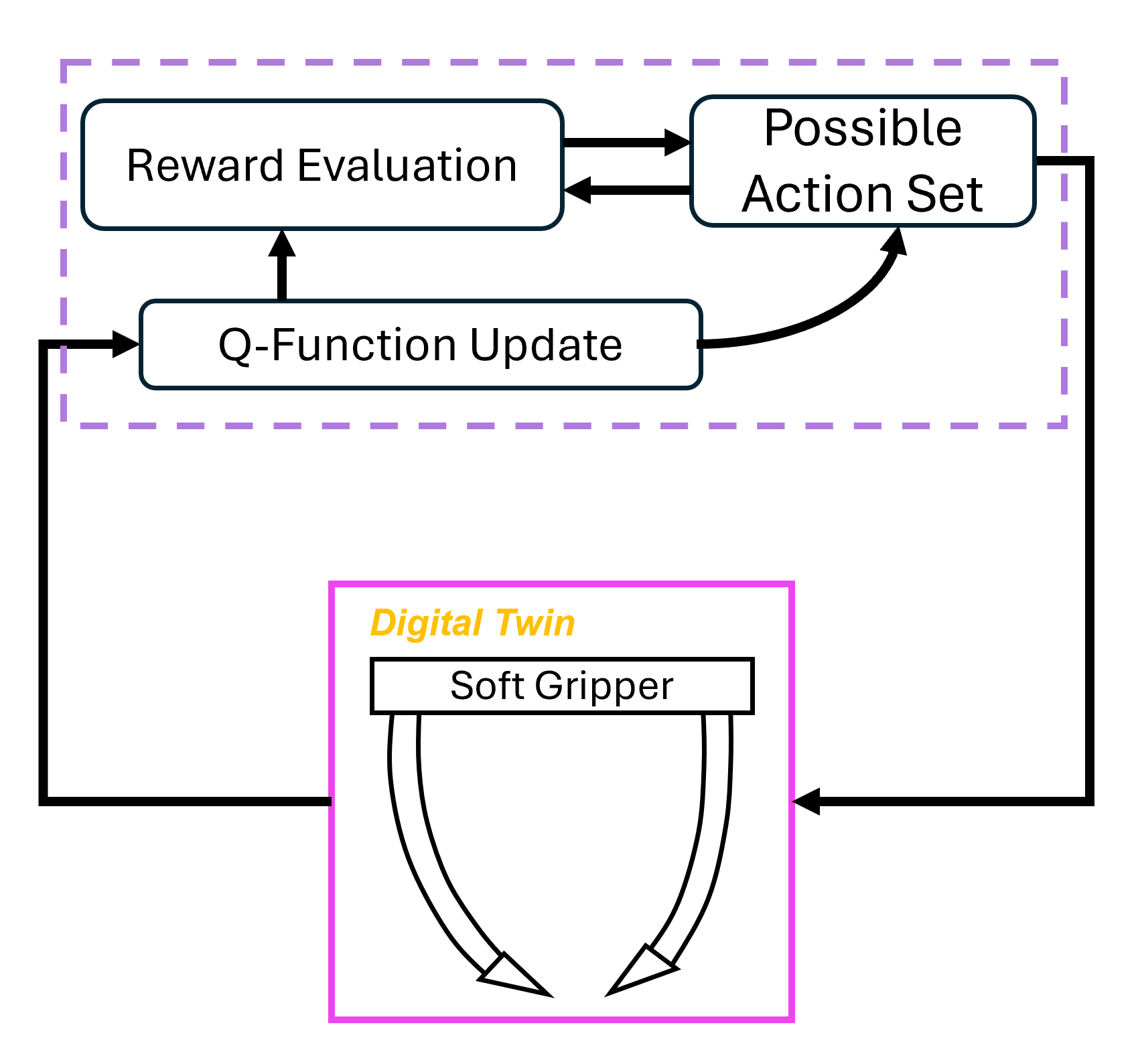}
    \caption{Q-learning is applied to determine the optimal motion speed for the digital twin of the soft gripper in order to achieve underactuated motions.}
    \label{fig_1}
\end{figure}

Developing simulation environments for soft robots presents several challenges. Due to their elastic nature, soft robots undergo large shape transformations under load. Accurately modeling these materials is difficult because they deform in ways that are not easily captured by traditional rigid-body dynamics models. Additionally, soft materials exhibit uncertainty during deformation~\cite{rebecca2015material}, with extremely soft materials showing even greater unpredictability during operation. Other observed phenomena in soft materials include hysteresis and time-varying effects~\cite{wang2019id}. These factors complicate the development of simulation environments. However, as research progresses, the behavior of soft robots is gradually being deciphered, and roboticists' understanding of their properties continues to grow~\cite{rebecca2015material,Rebe2023material}. Recently, simulation environments for soft robots have been developed to support machine learning model training~\cite{graule2022simulator}.

This paper presents the development of a multi-fingered soft gripper simulator designed for the advancement of underactuation methods. Our goal is to create a referencable simulator for soft grippers to train machine learning models. During the design phase, we study and model key properties of soft materials within the simulator. These properties include nonlinearity, uncertainty, hysteresis, and time-varying effects—factors that have been widely acknowledged and verified in recent research. A referenable soft gripper digital twin enables effective sim-to-real skill transfer. Next, a reinforcement learning approach is employed to discover an underactuated control strategy, using a single input to control multiple soft fingers. Training the RL model within the digital twin successfully demonstrates its feasibility. Additionally, other machine learning algorithms can be applied to develop control strategies for this soft gripper configuration in the future.

To understand the contributions of this paper in a comparative manner, related works are discussed. Li et al.~\cite{li2022softro} proposed a deep reinforcement learning framework for vision-based soft robot control, which was simulated in the MuJoCo environment. However, the MoJoCo environment is not deformable, which limits the ability to model the properties of soft materials accurately. Arachchige et al.~\cite{arachchige2024softro} modeled and simulated a soft-limbed robots in the PyBullet environment, but the environment has limited support for deformable bodies. Graule et al.~\cite{graule2022simulator} developed the SoMoGym which provides a modular method to model soft robots in complex environments. As the best knowledge of authors, our simulator tries to consider properties of soft materials such as nonlinearity, uncertainty, hysteresis effect, etc. compared to the SoMoGym.
To conclude, we seek to develop a soft robot simulation environment which considers the complex physical properties of soft robots, so the simulator would be applied for advanced controls.

The remainder of this paper is organized as follows. Section 2 introduces the soft gripper simulator design. Section 3 discusses how reinforcement learning method is applied to discover underactuation approaches. Section 4 demonstrates the training and simulation results, and Section 5 discusses and concludes the work.

\section{Soft Gripper Simulator Design}
\label{simulator}
Robot simulators allow researchers, developers, and engineers to test algorithms and robot behaviors without the risk of damaging expensive hardware or causing safety hazards. This is especially important for autonomous systems or robots with AI algorithms working in unknown environments. Simulators enable faster testing of ideas and iteration of design, reducing the time and cost associated with building physical prototypes.

Recently, the advancement of physical engines such as PyBullet~\cite{Vatsal2022pybullet}, Gazebo~\cite{Qian2014Gazebo}, and MoJuco~\cite{Todorov2012Mujoco} has made robot simulators more accurate, realistic, and able to handle complex interactions, including soft robotics, multi-body dynamics, and friction. Those simulators drive faster testing of ideas especially train machine learning models.

As the development of simulators, they are used for soft robots which have complex properties such as nonlinearity, hysteresis effect, uncertainty, etc. Popular simulators for soft robots include Gazebo or MuJoCo; however, the environments are relatively ideal compared to the real-world soft robots. For example, the components in those existing physics engine are non-deformable. Recently, a few custom-made simulators such as SoMoGym~\cite{graule2022simulator} provide some benchmark tasks evaluations. A few real-world phenomena are not considered in the simplified environments. 

The simulator developed in this research focuses on the dynamics of a multi-fingered soft gripper for coordination control algorithm development. The properties and design concepts considered in this simulator will be discussed in the following sections.

\begin{figure}[t]
    \centering
    \includegraphics[width=220pt]{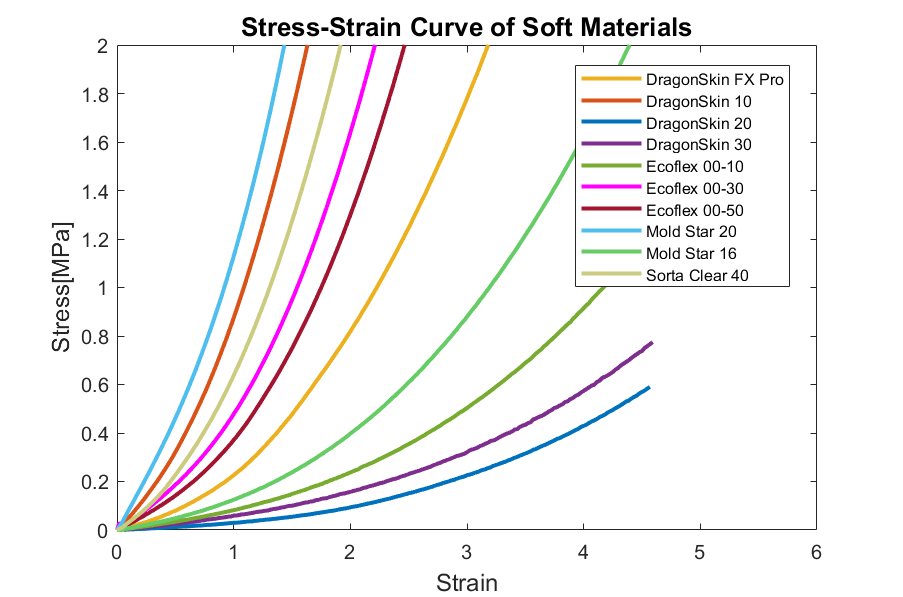}
    \caption{The stress-strain curve of soft materials demonstrate the nonlinearity of several commonly used materials in soft robots.~\cite{Marechal2021material}.}
    \label{fig_2}
\end{figure}

\subsection{Features of Soft Robots}
\label{feature}
Soft robots are made of soft materials such as polymers. Soft materials can undergo large deformation and recover to the initial state~\cite{kastor2017material} so they show different dynamics and properties compared to traditional rigid robots. Recent research~\cite{rebecca2015material, Rebe2023material} tried to decipher the properties of soft materials. Here, some properties will be discussed.

\subsubsection{Nonlinearity} A major feature of soft materials is the nonlinearity. Soft materials are much softer because they have much smaller Young's modulus~\cite{Tolley2018design}. They exhibit nonlinearity under large deformations as shown in Figure~\ref{fig_2}~\cite{Marechal2021material}. The more deformations they undergo, the higher nonlinearity they exhibit. However, soft robots show relatively linear properties under limited deformations~\cite{yang2024model}. That is, linear models catch the dynamics under limited deformations.

\subsubsection{Hysteresis Effect} The hysteresis effect is influenced by the softness of the materials. Softer materials, with lower Young's modulus, exhibit a more pronounced hysteresis effect~\cite{rebecca2015material}. One way to address this is by using distinct models for the loading and unloading paths. Namely, the model parameters for soft robots should differ between the loading and unloading phases as shown in Figure~\ref{fig_3}.

\subsubsection{Uncertainty} The commonly used soft materials, especially \textit{Smooth-on Ecoflex Series}, show higher uncertainty when they are undergoing slow deformations rates~\cite{rebecca2015material}. Reversely, their uncertainty reduces when higher deformation rates are applied. This property influences the dynamics and control of soft robots including the multi-fingered soft gripper~\cite{yang2024control}. Slow motions amplify uncertainty, while fast motions reduce it.

\subsubsection{Time-Varying Effect} Soft matters change behaviors over time by the Mullins effect~\cite{mullins1969soften}, which is important for modeling and understanding the dynamics of soft robots. When soft robots do repetitive works, they tend to become softer and their dynamics change~\cite{rebecca2015material}. For example, if soft grippers pick and place an object repeatedly, the time-varying effect would influence the grasping accuracy and should be considered to ensure the success of grasping.

\begin{figure}[t]
    \centering
    \includegraphics[width=210pt]{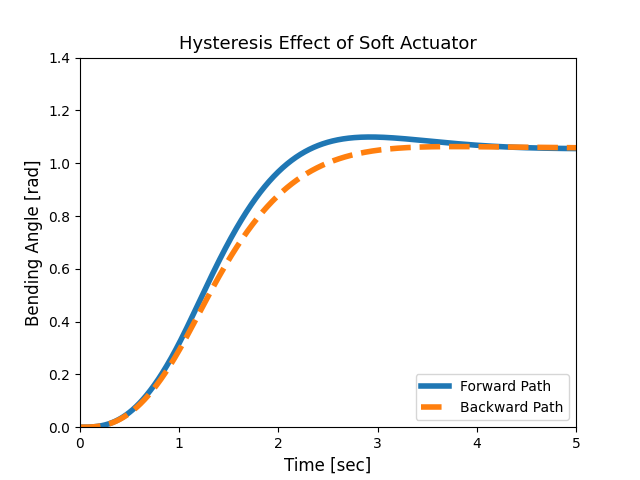}
    \caption{The hysteresis effect differs the forward ($\zeta = 0.7$) and backward ($\zeta = 0.8$) responses of a soft actuator in the simulator.}
    \label{fig_3}
\end{figure}

\subsection{Simulator Design}
\label{simulator}
The objective of building the simulator is to make it closely resemble real-world robots. As the simulator reflects the real scenarios, the simulation results can be referenced and will be more implementable. This research will design a soft robot simulator which considers the features discussed in Sec.~\ref{feature}.  

The soft grippers in this simulator are driven by soft pneumatic actuators, and the dynamics of soft pneumatic actuators is described as

\begin{align}
    \begin{split}
    {M_{eq}} \ddot{\theta} + C_n{\dot \theta} + K_n {\theta} = F(p)
    \label{eqn_1}
    \end{split}
\end{align}
where $M_{eq}$ is the equivalent mass of the soft actuator and $C_n$ is the damper of the soft actuator. The $M_{eq}$ and $C_n ($or $\zeta)$ currently are estimated by system identification through MATLAB\textregistered. The $F(p)=A\cdot p$ is the equivalent force generated by the air pump with pressure $p$ and $A$ is a constant determined by experimentation. $K_n$ is the spring constant and is represented as

\begin{align}
    \begin{split}
    K_n = \frac{2EI}{L^2}
    \label{eqn_1_1}
    \end{split}
\end{align}
where $E$ is the Young's modulus, $I$ is the moment of inertia, and $L$ is the length of the soft actuator~\cite{yang2023control}.

The hysteresis effect of the soft pneumatic actuator is affected by the damping constant $C_n$ as shown in Figure~\ref{fig_3}. The forward and backward paths have a different damping constant. The forward path has a smaller damping constant while the backward path has a larger damping constant.

The time-varying effect of the soft pneumatic actuator is offset by the natural frequency or the $K_n$ in (\ref{eqn_1}). As the soft gripper repeats pick-and-place tasks, the $K_n$ of the soft actuator will gradually reduce. Therefore, the $K_n$ in (\ref{eqn_1}) decreases as the cyclic motions increase.

The soft pneumatic actuator is powered by an air pump, which is driven by a linear motor~\cite{yang2023pump}. The dynamic equation of an air pump is represented as
\begin{align}
    \begin{split}
    \dot p = B\cdot \omega
    \label{eqn_1_2}
    \end{split}
\end{align}
where $B$ is a constant determined by the specifications of the air pump, and $\omega$ is the motor speed. 

These dynamical equations introduce in this section are programmed in the Python environment to build the digital twin of soft robotic grippers.

\begin{figure}[t]
    \centering
    \includegraphics[width=210pt]{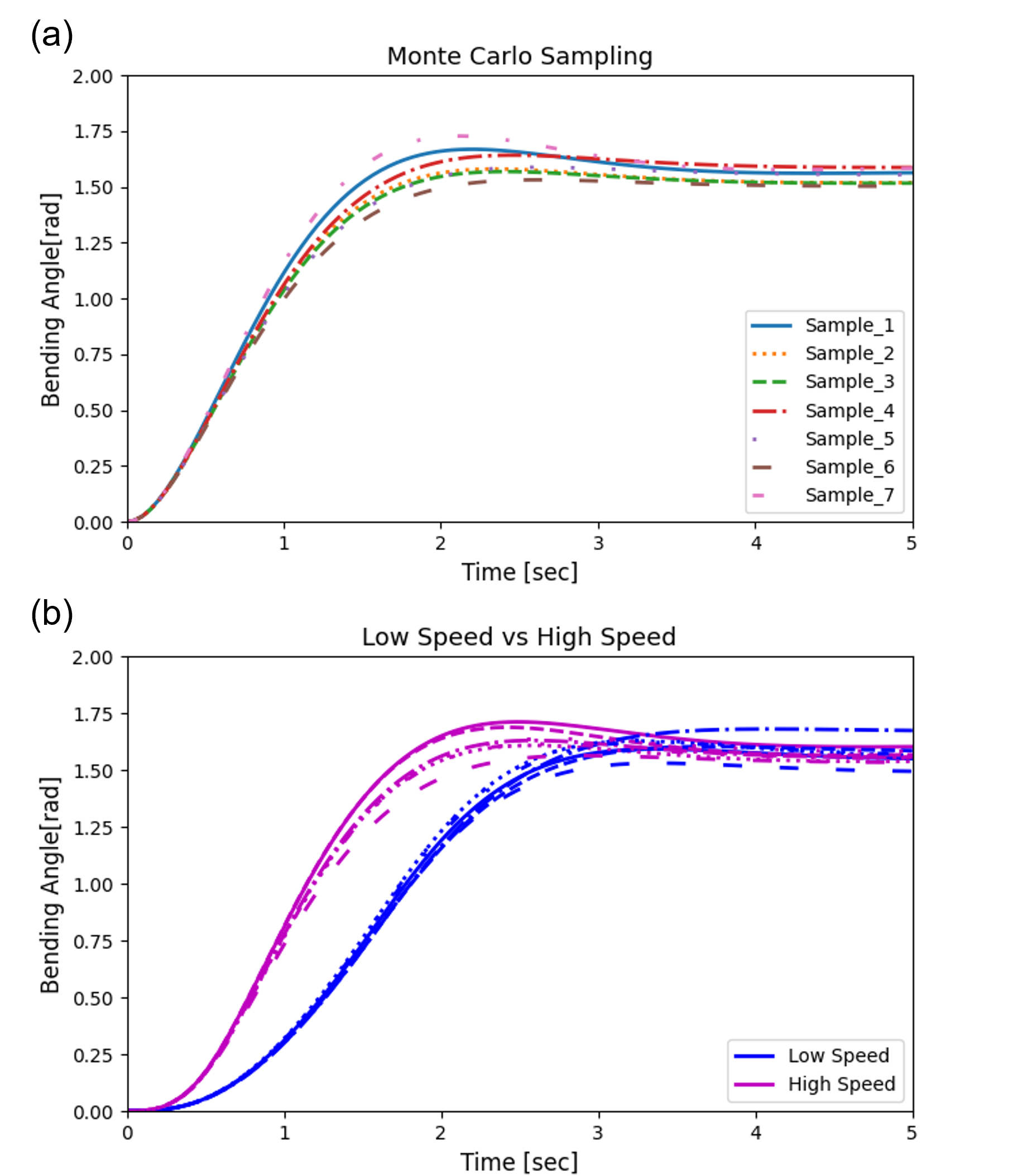}
    \caption{(a) Monte Carlo method is applied to simulate the dynamics of a soft pneumatic actuator under uncertainty. (b) High speed and low speed responses of soft pneumatic actuator and their excited uncertainty. The standard deviation of the steady-state errors for low-speed motions is around 0.1 $rad$ while that of high-speed motions is around 0.05 $rad$.}
    \label{fig_4}
    \vspace{-0.1in}
\end{figure}

\subsection{Monte Carlo Method}
It is observed that soft materials exhibit uncertainty during cyclic motions~\cite{rebecca2015material, yang2024control}. The uncertainty of soft matters is dependent on its motion speeds. Slow speed motions induce high uncertainty while high speed motions reduce it. The uncertainty varies with the damping ratio and the natural frequency~\cite{yang2024model}. Here, the damping ratio and the natural frequency are assumed to have a Gaussian distribution and it is represented as
\begin{align}
    \begin{split}
    \zeta \sim \mathcal{N}({\zeta_{nom}}, \sigma_{\zeta}^2)\\
    {\omega_n} \sim \mathcal{N}({\omega_{n\_nom}}, \sigma_{\omega_n}^2)
    \label{eqn_2}
    \end{split}
\end{align}
where $\zeta_{nom}$ is the nominal value of ${\zeta}$, $\sigma_{\zeta}$ is the standard deviation of $\zeta$, ${\omega_{n\_nom}}$ is the nominal value of $\omega_n$ and $\sigma_{\omega_n}$ is the standard deviation of $\omega_n$.

Given the nominal value and standard deviation of $\zeta$ and $\omega_n$, Monte Carlo method~\cite{kroese2014mcm}, which uses random sampling, is used to simulate the uncertainty for soft pneumatic actuator as Figure~\ref{fig_4} (a). When user is running the simulation, the algorithm will randomly pick a transfer function within the variation interval to simulate the dynamics of soft fingers. The uncertainty effect of soft actuator is simulated by the Monte Carlo method. Since the uncertainty is found to be related to motion speeds as discussed in Sec.~\ref{feature}, the $\sigma_{\zeta}$ and $\sigma_{\omega_n}$ are proportional to a function of motions speed $\dot\theta$. 

\begin{align}
    \begin{split}
    \sigma_{\zeta}, \sigma_{\omega_n} \propto a(\dot\theta)
    \label{eqn_3}
    \end{split}
\end{align}
where ${\dot\theta} > 0$ is the bending speed of the soft pneumatic actuator, and $a(\dot\theta)$ is a monotonic function and a varies with soft materials~\cite{rebecca2015material}. The simulation results can be seen in Figure~\ref{fig_4} (b). The standard deviation of the steady-state error for low-speed motions is much higher compared to that for high-speed motions.

Lastly, even though the soft pneumatic actuators have exactly the same dimensions and are made of the same soft materials, their system parameter $\omega_n$ in (\ref{eqn_1}) and its standard deviation are different.  The Monte Carlo simulation is run for each soft pneumatic actuator to mimic the uncertainty in the simulator as the Sec.~\ref{setup} and \ref{traingRL}. So, we could get the digital twin of soft gripper which is closer to the real-world soft gripper.

\section{Underactuated Control Approaches}
\label{UA}
Soft pneumatic actuators are powered by air pumps, which are typically bulky in size. Underactuation occurs when a system has fewer actuators than degrees of freedom. Due to the growing need for size optimization, underactuated control strategies have become increasingly important. Yet underactuated control remains an open challenge~\cite{daniela2023survey}. This section explores the feasibility of using reinforcement learning for underactuated control of a soft gripper.  

\subsection{Problem Statement}
When a multi-fingered gripper is constructed, the system equation is formulated as following. We take Laplace transform of both (\ref{eqn_1}) and (\ref{eqn_1_1}). Since the soft pneumatic actuator is powered by the air pump, both equations are cascaded as

\begin{align}
    \begin{split}
    TF = \frac{\bf{\Theta}}{\bf{P}}\times\frac{\bf{P}}{\bf{\Omega}} = \frac{C}{M_{eq}s^3 + C_n s^2 + K_n s}
    \label{eqn_1_3}
    \end{split}
\end{align}
where $C = A\cdot B$. Thus, the underactuated soft gripper is represented as 

\begin{align}
\begin{bmatrix} Y_{1}(s)\\ \vdots \\ Y_{n}(s) \end{bmatrix} &= \begin{bmatrix} P_1(s)\\ \vdots \\ P_{n}(s) \end{bmatrix} {U(s)} \\
\Rightarrow Y(s) &= P(s)U(s)
\label{eqn_4}
\end{align}
where $n$ is the number of finger, $P_i(s) \in \Bbb{R}(s)$, $Y_i(s) \in \Bbb{R}(s)$, $i = 1 ... n$, $U(s) \in \Bbb{R}(s)$, and $\Bbb{R}(s)$ is the set of rational number over $s$. $P_i(s)$ is the (\ref{eqn_1_3}). Since soft actuators have different system parameters such as $C_n$ and $K_n$, each $P_i(s)$ is generally not equal (i.e., $P_1(s) \neq P_2(s) \neq ... \neq P_n(s)$). $Y_i(s)$ is desired output function of $P_i(s)$ and it is step function in this study.

\subsection{Reinforcement Learning}
\label{RL}
The underactuation of multi-fingered soft gripper is achievable under some certain assumptions and conditions as discussed in Sec.~\ref{introduction}. The experimental results support this approach~\cite{yang2024control}. However, the algebraic approach may not be effective when those conditions and assumptions are not valid. Since the growth and popularity of machine learning methods, machine learning-based control serves as an alternative approach for underactuated control. 

Due to the distinct models and model uncertainty, the coordination motions of soft grippers are hard to be achieved. As observed, the uncertainty in the soft gripper's performance is highly dependent on the actuation speed. At higher speeds, the uncertainty is reduced, leading to more reliable and precise gripping. However, the challenge lies in identifying the optimal speed that balances speed and uncertainty reduction, taking into account the dynamics of underactuation, where the single air pump controls multiple fingers with varying demands.

To address this, we use Q-learning to autonomously search for the optimal actuation speed that reduces the uncertainty in the gripper's performance. The system can be modeled as a Markov Decision Process (MDP), where the state $s_t$ is represented by the uncertainty level observed on each finger. At each time step $t$, the gripper takes an action $a_t$ which corresponds to adjusting the actuation speed~\cite{watkins1992qlearning}.

The Q-learning update rule for learning the action-value function $Q(s_t, a_t)$ is given by

\begin{small}
\begin{align}
Q(s_t, a_t) \leftarrow Q(s_t, a_t) + \alpha [r_t + \gamma \max_{a'} & {~Q(s_{t+1}, a') + Q(s_t, a_t)}]
\label{eqn_5}
\end{align}
\end{small}where $Q(s_t, a_t)$ represents the expected reduction in uncertainty for taking action $a_t$ at state $s_t$, $\alpha$ is the learning rate, $\gamma$ is the discount factor, $r_t$ is the immediate reward, and $max_{a'} Q(s_{s+1} , a')$ is the maximum expected reduction in uncertainty for the next state $s_{t+1}$.

The reward $r_t$ is defined based on the reduction in uncertainty after adjusting the actuation speed. The goal is to minimize uncertainty, so the reward function is designed to increased as the uncertainty decreases:

\begin{align}
r_t = - \sigma(e_{ss}(s_{t+1}))
\label{eqn_6}
\end{align}
where $\sigma$ is the standard deviation of steady-state error for soft actuator at $s_{s}$. Over time, the gripper learns to select the optimal speed $a_t$ that minimizes uncertainty by updating the Q-values and gradually converging the optimal policy.

The agent (soft gripper) explores different speed settings and updates a Q-table to represent the expected reduction in uncertainty for each state-action pair. Over time, the agent learns to select the speed that minimizes uncertainty effectively. This method allows for the optimization of the gripper's performance without the need for explicit testing and searching the optimal motion speed. By using Q-learning, we aim to achieve an optimal speed that reduces uncertainty in the soft gripper’s operation. The detail information on training the Q-learning algorithm is discussed in Sec.~\ref{traingRL}.

\section{Evaluation by Simulations}
\label{exp}
The simulation environment is designed and discussed in Sec.~\ref{simulator} and the underactuation approach is introduced in Sec.~\ref{UA}. Thus, this section attempts to evaluate the feasibility of the simulator and proposed underactuation strategy.

\subsection{Setup the Simulator}
\label{setup}
In this study, we consider $n=2$ to investigate the underactuation of a two-fingered gripper controlled by a single input as the Figure~\ref{fig_1}. The system parameters in (\ref{eqn_1}) - (\ref{eqn_1_2}) are obtained by referencing the existing research~\cite{yang2024control}. The parameters include the dimensions of the soft actuator and material properties. The damping ratio is 0.7 and the natural frequencies are 1.9 and 1.75 $rad/s$ for both fingers. Thus, the $C_n = 2\zeta\omega_n$ and $K_n = {\omega_n}^2$ are determined and $M_{eq}$ is around $0.18~N$. With all the parameters, the soft gripper is initialized in the simulation environment and is ready for machine learning model training.

\begin{figure}[t]
    \centering
    \includegraphics[width=210pt]{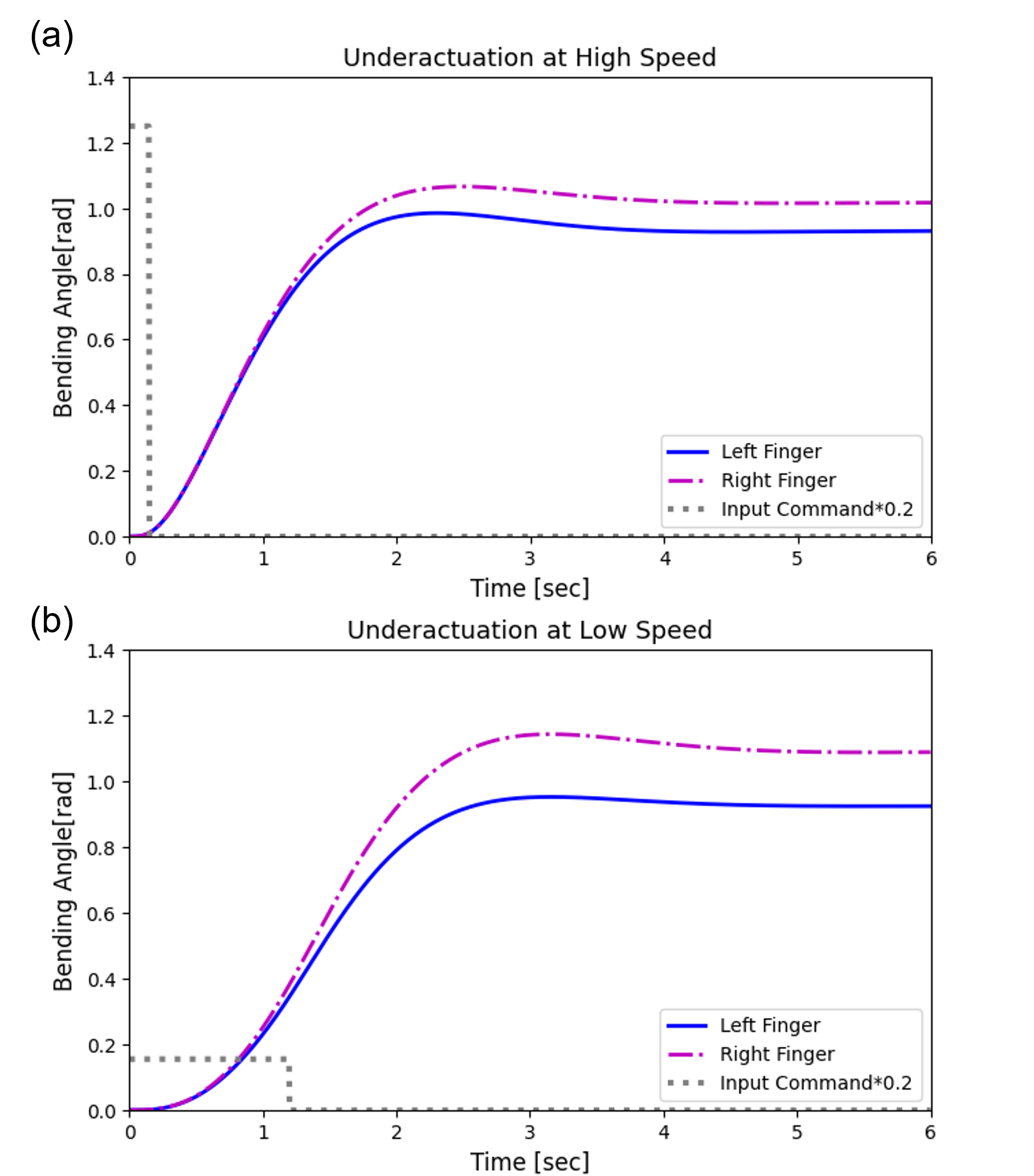}
    \caption{(a) Simulation of underactuation in the two-fingere soft gripper at high speed. (b) Simulation of underactuation in the two-fingered soft gripper at low speed.}
    \label{fig_5}
    \vspace{-0.1in}
\end{figure}

\subsection{Training RL Model}
\label{traingRL}
The soft gripper is powered by an air pump driven by a linear motor, as discussed in Sec.~\ref{simulator}. The linear motor's operating speed in the air pump ranges from $0 - 2\pi~rad/s$, which is divided into six speed intervals: $[\pi/3, 2\pi/3, \pi, 4\pi/3, 5\pi/3, 2\pi]~rad/s$. Based on empirical results, the corresponding uncertainty pair of $\zeta$ and $\omega_n$ ($\sigma_{\zeta}, \sigma_{\omega_n}$), representing the standard deviations of the $\zeta$ and $\omega_n$ in the soft actuator, are $[(0.12, 0.1), (0.1, 0.084), (0.08, 0.071), (0.06, 0.063),\\ (0.052, 0.055), (0.05, 0.055)]$. The steady-state errors could be estimated by applying final value theorem to (\ref{eqn_1_3}).

By implementing the Q-learning algorithm, it will search for the optimal speed within the defined ranges to minimize the steady-state errors in the soft actuators of the soft gripper. The $\alpha$ and $\gamma$ in (\ref{eqn_5}) are set as $0.1$ and $0.95$ respectively. Based on the uncertainty parameters ($\sigma_{\zeta}, \sigma_{\omega_n}$), the speeds such as $5\pi/3$ and $2\pi$ $rad/s$ will be regularly visited, which have better rewards $r_t$. The Q-learning training results confirm this expectation. Over 10 episodes, the algorithm finds the optimal speed that reduces the uncertainty. 


\subsection{Multi-Fingered Gripper Coordination}
The simulator runs the soft gripper based on the searching results. The soft actuators in the soft gripper are controlled at high speed (speed of linear motor $2\pi~rad/s$). The results are shown in the Figure~\ref{fig_5} (a) and both fingers are coordinated and their differences are below 5 $deg$ in the transient state. By contrast, if they are operated at low speed (speed of linear motor $\pi/12~rad/s$), both fingers could differ above $10~deg$ during the transient and steady states. The differences will lead to grasping failure in applications. 

In addition, the input commands are the gray dashed lines in both Figure~\ref{fig_5} (a) and (b). They are constant values in this research. The input commands could vary and generate more smooth motions for the soft gripper if advanced RL algorithms are applied to search the optimal input commands.

\section{Discussion and Conclusion}

\subsection{Discussions}
Certain parameters must be determined to initialize the soft gripper digital twin for machine learning model training. The system parameters in (\ref{eqn_1_1}) are influenced by materials properties ($E$) and dimensions of soft actuators ($I$ and $L$). Using $E$, $I$, and $L$, the spring constant $K_n$ can be computed. The damping ratio ($\zeta$) and equivalent mass ($M_{eq}$) in (\ref{eqn_1}) are obtained through system identification by analyzing the responses of soft actuators in MATLAB\textregistered. Note that $C_n$ will differ between the forward and backward paths as discussed in Sec.~\ref{feature}, so it should be identified separately. In addition, the $a(\dot\theta)$ and $\sigma_{\zeta}$ and $\sigma_{\omega_n}$ in (\ref{eqn_3}) are identified by testing soft actuators at high and low speeds. Once these parameters are determined, the digital twin in the simulator is ready to train machine learning models. 

The problem is simplified for the reinforcement learning model, Q-learning, which traditionally operates on discrete states and actions. In the context of a soft gripper, however, the motor speed and other system parameters (e.g., finger angular velocity, uncertainty levels) are continuous variables in real applications.  An alternative approach involves using function approximation techniques, such as Deep Q-learning (DQN), which approximates the Q-values using a neural network. This method can handle continuous state spaces by mapping continuous values to approximate Q-values, allowing the agent to generalize across a wider range of states.

\subsection{Conclusion}

This study presents the design of a multi-fingered soft gripper digital twin that accounts for commonly observed phenomena such as nonlinearity, hysteresis, time-varying behavior, and uncertainty. The dynamics of the soft actuators in the gripper are modeled by a linear second-order system within the operation range of $[0, 8\pi/9]~rad$. These phenomena are represented by variations in the damping and spring constants. Notably, uncertainty of soft actuator is simulated using Monte Carlo sampling, based on the standard deviations of the damping and spring constants. Once the simulator is constructed, a Q-learning algorithm is applied to identify the optimal speed that minimizes the uncertainty in a two-fingered soft gripper, thereby enabling coordinated motions in the simulations. This digital twin enables the simulation of soft gripper dynamics and the development of advanced control algorithms.

\subsection{Future Works}
Future work will extend this to model the time-varying effects of soft grippers to complete this simulator. Also, we would incorporate contact features, specifically examining the interactions between the fingertips and target objects based on contact mechanics theories~\cite{Kenneth1987mechanics}. Furthermore, the sim-to-real skill transfer will be applied to real-world soft grippers. As a result, the digital twin will be capable of predicting the grasping success rate before execution.





\section*{ACKNOWLEDGMENT}
This work is supported by the National Science and Technology Council, Taiwan, under contract: MOST 113-2634-F-007-002 -.

\bibliographystyle{ieeetr}
\bibliography{IEEEabrv}

\end{document}